\documentclass{article}

 \usepackage[preprint]{neurips_2026}

\usepackage[utf8]{inputenc} 
\usepackage[T1]{fontenc}    
\usepackage{hyperref}       
\usepackage{url}            
\usepackage{booktabs}       
\usepackage{amsfonts}       
\usepackage{nicefrac}       
\usepackage{microtype}      
\usepackage{xcolor}         
\usepackage{amsmath}         
\hypersetup{
    colorlinks,
    linkcolor={red!50!black},
    citecolor={blue!50!black},
    urlcolor={blue!80!black}
}

\usepackage{graphicx} 
\usepackage{threeparttable}
\usepackage{subcaption}
\usepackage{algorithm}
\usepackage{algpseudocode}
\usepackage{enumitem}
\title{DuSPiT: Dual-Branch Sub-Patch Pixel Diffusion Transformer }

\author{%
  Yunpeng Bai \\
  The University of Texas at Austin\\
  \And
  Yossi Gandelsman \\
  Reve\\
  \And
  Micha\"el Gharbi \\
  Reve\\
}

\begin{document}

\maketitle


\begin{abstract}

Diffusion Transformers achieve strong image generation performance, but most operate in compressed latent spaces. Pixel-space diffusion avoids this information loss, yet existing approaches map each raw image patch to a single token, forcing one representation to handle both global communication and fine-grained details.
We address this issue by proposing a new architecture, \textbf{DuSPiT}, a \textbf{Du}al-branch \textbf{S}ub\textbf{P}atch \textbf{Pi}xel \textbf{T}ransformer. This model separates global structural reasoning from local appearance modeling. DuSPiT uses a compact base branch for efficient global reasoning and a parallel, high-capacity pixel branch, organized into subpatch groups, to preserve detailed appearance, with the two branches interacting through cross-attention. Our results show that DuSPiT generates images with richer details and stronger fine-grained structures, while also achieving a better quality--efficiency trade-off than prior pixel-space diffusion transformers. The code is available at: \url{https://github.com/yunpeng1998/DuSPiT}.

\end{abstract}


\section{Introduction}

Diffusion Transformers (DiTs)~\cite{dit} have become a dominant paradigm for high-quality image generation, and most modern systems improve efficiency by performing diffusion in the latent space of a pretrained variational autoencoder~\cite{LDM}. However, this efficiency comes from substantial learned compression, which can discard fine-grained visual signals. This motivates renewed interest in pixel-space diffusion, where generation is performed closer to the raw image domain.

Pixel-space diffusion transformers have recently attracted growing interest~\cite{jit,yu2025pixeldit,ma2025deco}, as they operate directly on raw image patches and avoid the information loss introduced by learned latent compression. This makes them a promising direction for preserving high-frequency details such as textures, sharp boundaries, and textual patterns. However, it also creates a structural modeling challenge: each patch token must support global image-level interactions through attention while also preserving rich intra-patch structure for fine-grained appearance modeling. Existing methods typically compress each raw patch into a single compact token before applying standard transformer blocks, forcing one representation to support both long-range patch-level reasoning and fine-grained appearance modeling. While computationally efficient, this design can limit representation quality, and simply increasing token width quickly incurs substantial parameter and compute overhead.

In this work, we argue that the central challenge of pixel-space diffusion is not simply insufficient model width, but the absence of an architectural separation between \emph{efficient global patch-level reasoning} and \emph{high-fidelity local subpatch-level appearance modeling}. To address this issue, we propose \textbf{DuSPiT}, a \textbf{Du}al-branch \textbf{S}ub\textbf{P}atch \textbf{Pi}xel \textbf{T}ransformer. Instead of collapsing each raw patch into a single compressed token, DuSPiT maintains two coupled representations for every patch: a compact \emph{base branch} for efficient global interaction, and a parallel high-capacity \emph{pixel branch} that preserves patch appearance without undergoing the same aggressive compression.

Crucially, the pixel branch is not modeled as an unstructured high-dimensional vector. Rather, it is organized into multiple \emph{subpatch groups}, each corresponding to a finer spatial region inside the patch. This design preserves the internal spatial organization of raw pixel content and enables the model to represent intra-patch appearance at a much finer granularity. The base branch provides globally contextualized structural and semantic information, while the subpatch-aware pixel branch focuses on detailed appearance refinement. The two branches interact through cross-attention, allowing the model to combine efficient global reasoning with high-capacity local detail modeling, without forcing the entire transformer to operate in a prohibitively high-dimensional space.

To align these two levels of representation, we further introduce a hierarchical positional encoding scheme that combines patch-level positions with relative subpatch offsets. This gives the model a coherent notion of both inter-patch layout and intra-patch geometry, enabling each subpatch representation to attend to globally meaningful patch features while retaining precise local spatial identity.

Finally, DuSPiT predicts the clean image directly from the pixel branch, while the base branch serves purely as an internal structural bottleneck. This explicit division of labor is central to our design: global reasoning is carried out in a compact communication space, whereas high-fidelity reconstruction is preserved in a subpatch-aware appearance space.

Extensive experiments show that DuSPiT improves visual fidelity in pixel-space diffusion, especially for fine textures and sharp structures, while maintaining competitive efficiency. More broadly, our results suggest that effective pixel-space diffusion requires architectures that explicitly respect the dual role of image patches: they are both global communication units and structured containers of local appearance.

Our main contributions are summarized as follows:
\begin{itemize}[leftmargin=*]
    \item We address a structural quality issue in existing pixel-space diffusion transformers, where compact patch tokens must simultaneously support global patch-level interactions and fine-grained intra-patch appearance modeling.
    
    \item We propose \textbf{DuSPiT}, a dual-branch subpatch-aware pixel diffusion transformer that maintains a compact base branch for efficient global reasoning and a parallel high-capacity pixel branch for detailed appearance modeling.
    
    \item We introduce a hierarchical positional encoding scheme that aligns patch-level structure with subpatch-level spatial detail, enabling effective cross-branch interaction.
    
    \item We show that DuSPiT improves pixel-space image generation quality, particularly on fine textures, while remaining computationally efficient.
\end{itemize}
\section{Related Work}

\subsection{Image Generation}

Recent state-of-the-art image generation systems are largely built on latent diffusion models (LDMs~\cite{LDM}), which perform denoising in a compressed latent space. This design greatly reduces computation and memory cost, enabling larger backbones and higher-resolution generation. In practice, LDMs typically rely on a VAE tokenizer to balance reconstruction fidelity and compression efficiency. Many recent works therefore focus on improving the autoencoding stage through better architectures, objectives, and tokenization strategies~\cite{dcae,dcae1p5,maetok,yu2025zipir}, while also studying the trade-off between reconstruction quality and generative performance~\cite{lgt}. Beyond these improvements, several studies explore end-to-end optimization of the latent representation. For instance, REPA-E jointly trains the autoencoder and diffusion transformer to better align the learned latent space with the generative objective~\cite{repa-e}. More recently, alternative formulations have replaced the traditional variational bottleneck with representation autoencoders, demonstrating competitive latent representations without relying on explicit variational modeling~\cite{shi2025svg,zheng2025rae}. Collectively, these advances have established latent diffusion as the dominant framework for contemporary high-quality image synthesis.

\subsection{Pixel-Space Diffusion Models}

Diffusion models operating directly in pixel space predate latent-based approaches and continue to attract significant research interest. Early works showed that directly denoising images in the pixel domain can produce highly realistic results~\cite{adm}. Subsequent developments, such as cascaded diffusion models, extended this capability to higher resolutions by employing multi-stage, coarse-to-fine synthesis pipelines~\cite{cdm}. Despite these successes, a major limitation of pixel-based approaches is their unfavorable scalability: both computational and memory costs grow quadratically with image resolution, making end-to-end training at megapixel scales particularly challenging. More recent efforts~\cite{epg,farmer,jetformer,li2025fractal} have revisited pixel-space generation with innovations in both architecture and training methodology. Simple Diffusion improves efficiency through streamlined convolutional networks equipped with skip connections~\cite{simplediffusion,sid2}. PixelFlow introduces a hierarchical flow-based framework for modeling pixel-level distributions~\cite{pixelflow}. PixNerd further enhances computational efficiency by integrating lightweight neural field layers into the diffusion process~\cite{pixnerd}. Notably, JiT~\cite{jit} demonstrates that standard Transformer architectures can effectively operate in pixel space by directly predicting clean images using an $x_0$-prediction objective. Concurrent works such as PixelDiT~\cite{yu2025pixeldit} and DeCo~\cite{ma2025deco} follow a similar paradigm, where a patch-level DiT first generates coarse structural representations, which are then refined into full-resolution pixel-space images via a pixel-level decoder.


\begin{figure}
    \centering
    \includegraphics[width=\linewidth]{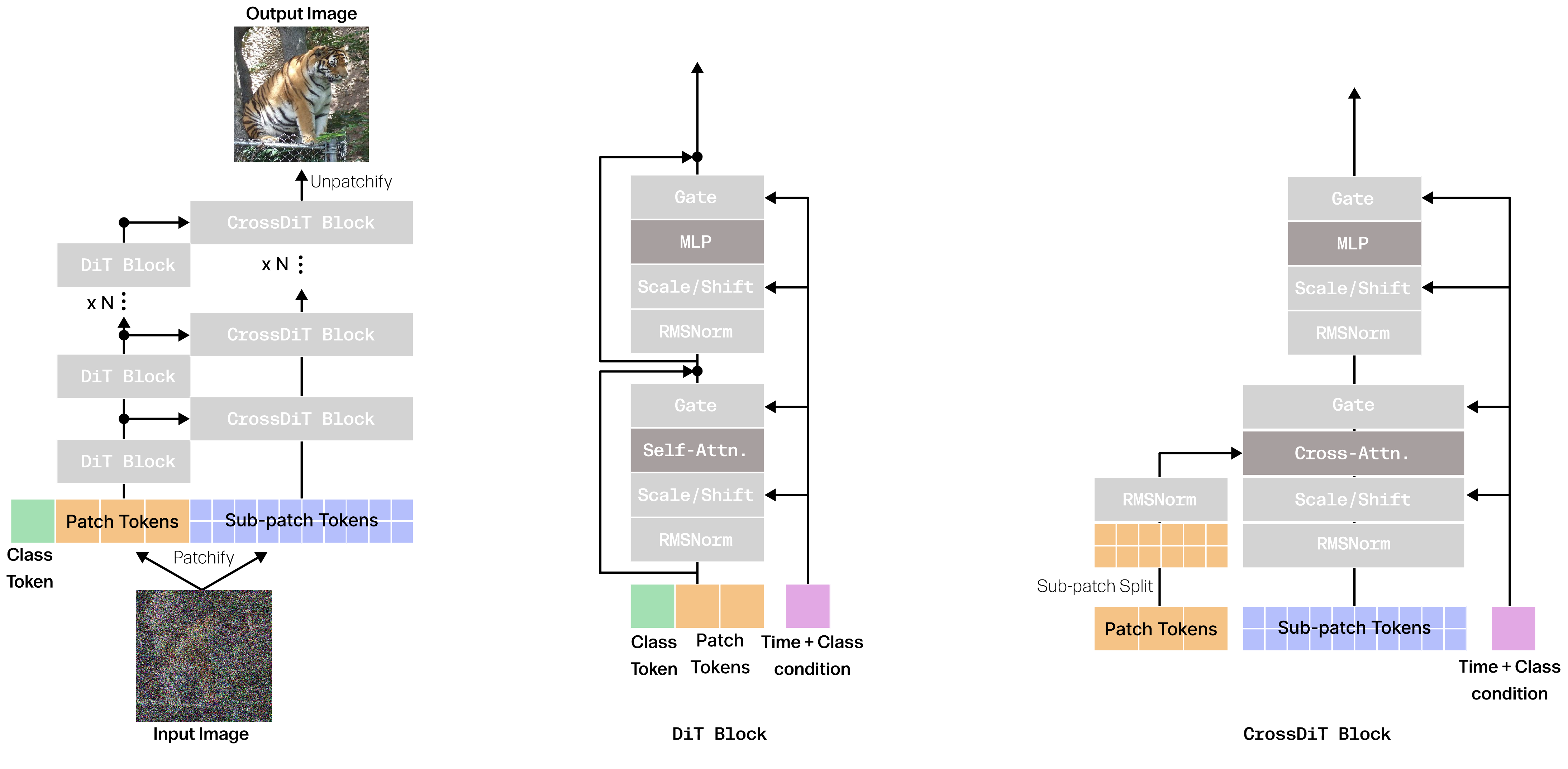}
    \caption{\textbf{DuSPiT architecture.} (a) The model takes a noisy input and processes it with two patch embedding branches: a coarse token branch and a fine sub-token branch. Timestep and class embeddings are combined into a shared conditioning vector, which modulates all transformer layers. The coarse branch is updated by stacked DiT blocks, while its intermediate features are reshaped and used to guide the fine branch through CrossDiTBlock layers. The final fine-branch features are then decoded by a final layer and unpatchified to produce the output. (b) Structure of a DiT block with AdaLN-based modulation, self-attention, and an MLP. (c) Structure of the CrossDiTBlock, where coarse-branch features interact with fine-branch tokens through cross-attention followed by a token-wise SwiGLU feed-forward module.}
    \label{fig:arch}
\end{figure}
\section{Dual-Branch Sub-Patch Pixel Diffusion Transformer}

We propose a pixel space diffusion transformer architecture with an additional high-dimensional branch to better model fine-grained appearance details. The key idea is to decouple \textit{global patch-level structure} from \textit{local appearance modeling}. Therefore, our model contains a base branch that processes compact patch tokens for efficient global reasoning, and a pixel branch that maintains richer per-patch features for preserving appearance details. The two branches interact through cross-attention, allowing the patch branch to provide structural guidance while the pixel branch retains fine-grained information. Within each block, the model performs patch-wise attention over patch tokens, followed by intra-patch refinement of pixel tokens conditioned on the patch-level representation. Finally, the denoising prediction is generated solely from the pixel branch through a lightweight output head.

\subsection{Tokenization and Dual-Branch Representation}

Pixel-space diffusion transformers typically tokenize an image by splitting it into non-overlapping patches. Let $x_t \in \mathbb{R}^{H \times W \times C}$ denote the noisy image at diffusion step $t\in[0,1]$. The image is divided into patches of size $P \times P$, resulting in $N = HW / P^2$ patches. Each patch contains $P^2 \times C$ raw pixel values. For example, when $P=32$ and $C=3$, each patch corresponds to a 3072-dimensional vector. 
Representing such high-dimensional patch content is non-trivial. Unlike latent diffusion models, which rely on a pretrained VAE to compress images into a compact latent space, pixel-space diffusion lacks an explicit compression mechanism. Directly feeding tokens of dimension $P^2C$ into a transformer would lead to prohibitively large parameter counts and computational costs, especially since the complexity of attention and feed-forward layers scales with the embedding dimension.

Consequently, existing Pixel Diffusion Transformer (DiT) architectures typically project each raw patch into a lower-dimensional embedding using a simple linear, MLP, or convolutional layer:
\begin{equation}
z_i = E_{\text{patch}}(x_t^{(i)}), \qquad z_i \in \mathbb{R}^{d_p}, \quad d_p \ll P^2C,
\end{equation}
where $x_t^{(i)}$ denotes the $i$-th image patch. While this approach enables efficient transformer computation, the projection inevitably discards fine-grained information. In contrast to VAEs, which are explicitly trained to learn an information-preserving compression, such shallow projection layers lack sufficient capacity to achieve comparable reconstruction fidelity, often resulting in the loss of subtle textures and textual details.

\paragraph{Motivation for High-Dimensional Appearance Modeling.}

We argue that maintaining a representation with dimensionality comparable to, or even larger than, the original patch dimension is essential for accurately modeling appearance in pixel-space diffusion. However, naively increasing the transformer embedding dimension would significantly inflate the number of parameters and the computational cost of self-attention and feed-forward networks. To overcome these limitations, we introduce a \textit{dual-branch representation} that separates the roles of global communication and local appearance modeling:

\paragraph{Patch Branch.}
We define a compact patch token:
$
z_i \in \mathbb{R}^{d_p},$
where $d_p$ is a relatively small embedding dimension. This branch functions similarly to the standard DiT architecture and is responsible for capturing global spatial structure and semantic relationships across patches.

\paragraph{Pixel Branch.}
In parallel, we introduce a higher-capacity pixel token:
$
a_i \in \mathbb{R}^{d_a}, \quad d_a \geq P^2C \ \text{and} \ d_a \gg d_p,
$
which preserves rich appearance information such as textures, colors, and fine structural details. This branch is designed to retain the expressive power of raw pixel representations without forcing the entire transformer to operate at this high dimensionality.

The two representations are obtained via separate embedding functions:
\begin{equation}
z_i = E_{\text{patch}}(x_t^{(i)}), \qquad
a_i = E_{\text{pixel}}(x_t^{(i)}),
\end{equation}
where $E_{\text{patch}}(\cdot)$ can be implemented as a convolutional or linear projection similar to standard DiT models, and $E_{\text{pixel}}(\cdot)$ maps the raw patch to a higher-dimensional appearance space.

This dual-branch design enables the model to maintain high-fidelity appearance representations while keeping the computational cost of global reasoning manageable.

\paragraph{Cross-Branch Interaction.}
To effectively integrate global structure and local appearance, we enable information exchange between the two branches through cross-attention mechanisms. The patch tokens provide globally contextualized structural guidance, while the pixel tokens focus on detailed appearance modeling. This structured interaction allows the model to achieve a favorable balance between representation capacity and computational efficiency.

\subsection{Hierarchical Positional Encoding for Cross-Attention}

To effectively bridge the compact patch branch and the high-capacity pixel branch, we design a hierarchical positional encoding (PE) scheme for the cross-attention mechanism. Our goal is to preserve the global spatial consistency provided by standard Diffusion Transformers (DiTs) while enabling fine-grained appearance modeling within each patch.

\paragraph{Patch-Level Positional Encoding.}
In the base diffusion transformer, each patch token corresponds to the center of a spatial patch in the image. We adopt the standard 2D Rotary Positional Encoding (RoPE) \cite{su2024roformer} used in DiT to encode the global position of each patch. Let the spatial grid of patches be indexed by $(u, v)$ along the horizontal and vertical axes. The patch-level positional encoding is defined as:
\begin{equation}
\mathbf{p}_{\text{patch}}(u, v) = \mathrm{RoPE}(u, v),
\end{equation}
which is applied to the query, key, and value projections of the patch tokens during both self-attention and cross-attention.

\paragraph{Sub-Patch Appearance Encoding.}
While the patch-level encoding captures global spatial structure, it is insufficient for modeling fine-grained appearance details within each patch. To address this, we further partition the high-dimensional pixel token into a set of sub-patch groups that correspond to finer spatial regions inside each patch.

Specifically, for a patch of size $P \times P$, we conceptually divide it into an $n \times n$ grid of sub-patches. Each sub-patch represents a local region within the patch and is associated with a relative offset from the patch center. Let $(s, t)$ denote the sub-patch indices, where $s, t \in \{0, \ldots, n-1\}$. The relative offset of each sub-patch center is defined as:
\begin{equation}
\Delta_x(s) = \frac{s + 0.5}{n} - \frac{1}{2}, \qquad
\Delta_y(t) = \frac{t + 0.5}{n} - \frac{1}{2}.
\end{equation}

The final positional encoding for each sub-patch is obtained by augmenting the patch-level encoding with this relative offset:
\begin{equation}
\mathbf{p}_{\text{sub}}(u, v, s, t) = \mathrm{RoPE}\big(u + \Delta_x(s), \; v + \Delta_y(t)\big).
\end{equation}

This formulation enables the model to represent fine-grained spatial variations within each patch without explicitly encoding every individual pixel.

\paragraph{Dimension Partitioning of Pixel Tokens.}
Given a pixel token $a_i \in \mathbb{R}^{d_a}$, we partition it into $n^2$ groups corresponding to the sub-patch grid:
\begin{equation}
a_i = \left[ a_i^{(1)}, a_i^{(2)}, \ldots, a_i^{(n^2)} \right], \quad
a_i^{(r)} \in \mathbb{R}^{d_a / n^2}.
\end{equation}
For example, when $d_a = 3072$ and $n = 8$, the pixel token is divided into $8 \times 8 = 64$ sub-patch groups, each with dimension $48$. Each group $a_i^{(r)}$ is associated with a specific sub-patch location and is projected into the query space using a lightweight MLP. To reduce parameter overhead, these MLPs share parameters across all patches and sub-patch groups:
\begin{equation}
q_i^{(r)} = \mathrm{MLP}_q\!\left(a_i^{(r)}\right),
\end{equation}

\paragraph{Patch Tokens as Keys and Values.}

Let $z'_i \in \mathbb{R}^{d_p}$ denote the latent representation of the $i$-th patch token produced by the base diffusion transformer after the feed-forward (MLP) projection within each transformer block. These tokens encode globally contextualized structural information and serve as the keys and values in the cross-attention mechanism.

To align with the multi-head attention formulation and to facilitate efficient interaction with the sub-patch queries from the pixel branch, we partition $z'_i$ into $H$ attention heads:
\begin{equation}
z'_i = \left[ z'_i{}^{(1)}, z'_i{}^{(2)}, \ldots, z'_i{}^{(H)} \right], 
\qquad z'_i{}^{(h)} \in \mathbb{R}^{d_p / H},
\end{equation}
where $H$ denotes the number of attention heads.

For each head $h$, we compute the corresponding key and value vectors using head-specific linear projections:
\begin{equation}
k_i^{(h)} = W_k z'_i{}^{(h)}, 
\qquad 
v_i^{(h)} = W_v z'_i{}^{(h)},
\end{equation}
where $W_k$ and $W_v$ are learnable projection matrices.

\paragraph{Cross-Attention with RoPE.}
Rotary positional encoding (RoPE) is applied to the query and key vectors. Let $\mathcal{R}(\cdot, \mathbf{p})$ denote the RoPE operation. The rotated queries and keys are then given by:
\begin{equation}
\tilde{q}_i^{(r)} = \mathcal{R}\!\left(q_i^{(r)}, \mathbf{p}_{\text{sub}}(u, v, s, t)\right), \qquad
\tilde{k}_i^{(h)} = \mathcal{R}\!\left(k_i^{(h)}, \mathbf{p}_{\text{patch}}(u, v)\right).
\end{equation}

where the one-dimensional indices are obtained from their two-dimensional coordinates via row-major ordering:
$
i = u \cdot W_p + v,
r = s \cdot n + t,
$
with $W_p = W / P$. Here, $(u,v)$ denotes the patch coordinate, $(s,t)$ denotes the sub-patch coordinate within a patch, and $r$ is the flattened sub-patch index. In particular, $q_i^{(r)}$ denotes the query vector of the $r$-th sub-patch group within the $i$-th patch. During cross-attention, each sub-patch query $q_i^{(r)}$ from the pixel branch only attends to the corresponding patch token from the same patch:
\begin{equation}
\alpha_{i}^{(r,h)} =
\mathrm{softmax}_{h}\left(
\frac{ 
\tilde{q}_i^{(r)}
\left( \tilde{k}_i^{(h)} \right)^{\top} 
}
{\sqrt{d_h}}
\right), \qquad
\hat{a}_i^{(r)} = 
\sum_{h=1}^{H} \alpha_i^{(r,h)} v_i^{(h)} .
\end{equation}
where $d_h = d_p / H$ is the dimensionality of each attention head. All query, key, and value vectors are projected to the common dimension $d_h$ before computing attention.
Finally, the outputs from all heads and sub-patch groups within the same patch are concatenated to form the updated pixel representation:

\begin{equation}
\hat{a}_i = 
a_i +
\mathrm{Concat}_{r} \left( 
\mathrm{MLP}_{r}\!\left(\hat{a}_i^{(r)}\right)
\right).
\end{equation}
Here, each sub-patch group $r$ uses an independent projection $\mathrm{MLP}_{r}$, allowing different sub-patch locations to learn location-specific appearance transformations. As shown in Table~\ref{tab:ablation_modulation}, this design performs better than sharing a single MLP across all sub-patch groups.

\textbf{Output Head.}
After $L$ transformer blocks, the final denoising prediction is produced solely from the pixel branch. In contrast to the base patch branch, which mainly provides global structural guidance through cross-attention, the pixel branch preserves the high-capacity appearance representation used for reconstruction. Specifically, we directly project the final pixel tokens with a lightweight output layer $W_{\text{out}}$ to obtain the denoising prediction. We do not concatenate the patch tokens with the pixel tokens at the output stage. This design ensures that the final prediction is generated from the pixel branch alone, while the patch branch acts as a compact structural bottleneck for global reasoning. We adopt the $x_0$-prediction parameterization, where the model directly predicts the clean image rather than the noise term. The predicted pixel outputs are then rearranged back to the image space following the original patch layout to form the final reconstructed image $\hat{x}_0$.

\section{Experimental Results}

\begin{figure*}[t]
\centering

\begin{minipage}[t]{0.56\textwidth}
    \centering
    \small
    \setlength{\tabcolsep}{6pt}
    \begin{threeparttable}
    \captionof{table}{Comparison of latent-space and pixel-space diffusion models on ImageNet $512 \times 512$.}
    \label{tab:imagenet512_comparison}
    \begin{tabular}{lccc}
    \toprule
    \textbf{ImageNet $512 \times 512$} &
    \textbf{GFLOPs} &
    \textbf{FID}$\downarrow$ &
    \textbf{NFE} \\
    \midrule
    \multicolumn{4}{l}{\textit{Latent-space Diffusion}} \\
    DiT-XL/2~\cite{dit}                
        & 525 & 3.04 & 250 \\
    SiT-XL/2~\cite{sit}                
        & 525 & 2.62 & 250 \\
    REPA~\cite{repa}, SiT-XL/2         
        & 525 & 2.08 & 250 \\
    DDT-XL/2~\cite{ddt}                
        & 525 & \underline{1.28} & 250 \\
    RAE~\cite{zheng2025rae}, DiT$^{\text{DH}}$-XL/2
        & 642 & \textbf{1.13} & 50 \\
    \midrule
    \multicolumn{4}{l}{\textit{Pixel-space Diffusion}} \\
    ADM-G~\cite{adm}                   
        & 1983 & 7.72 & 250 \\
    RIN~\cite{rin}                     
        & 415  & 3.95 & 1000 \\
    SiD~\cite{sid}, UViT/4             
        & 555  & 3.02 & -- \\
    VDM++, UViT/4                      
        & 555  & 2.65 & -- \\
    SiD2~\cite{sid2}, UViT/4           
        & 137  & 2.19 & 512 \\
    SiD2~\cite{sid2}, UViT/2           
        & 653  & \underline{1.48} & 512 \\
    PixNerd~\cite{pixnerd}, XL/16      
        & 583  & 2.84 & 50 \\
    PixelDiT/32~\cite{yu2025pixeldit}  
        & 426  & 2.73 & 50 \\
    DeCo/16~\cite{ma2025deco}          
        & 508  & 2.22 & 50 \\
    JiT-G/32~\cite{jit}                
        & 384  & 1.78 & 50 \\
    \midrule
    DuSPiT/32                           
        & 329  & 1.52 & 50 \\
    DuSPiT/32                           
        & 329  & \textbf{1.46} & 512 \\
    \bottomrule
    \end{tabular}
    \end{threeparttable}
\end{minipage}
\hfill
\begin{minipage}[t]{0.40\textwidth}
    \vspace{15mm}
    \centering
    \includegraphics[width=\linewidth]{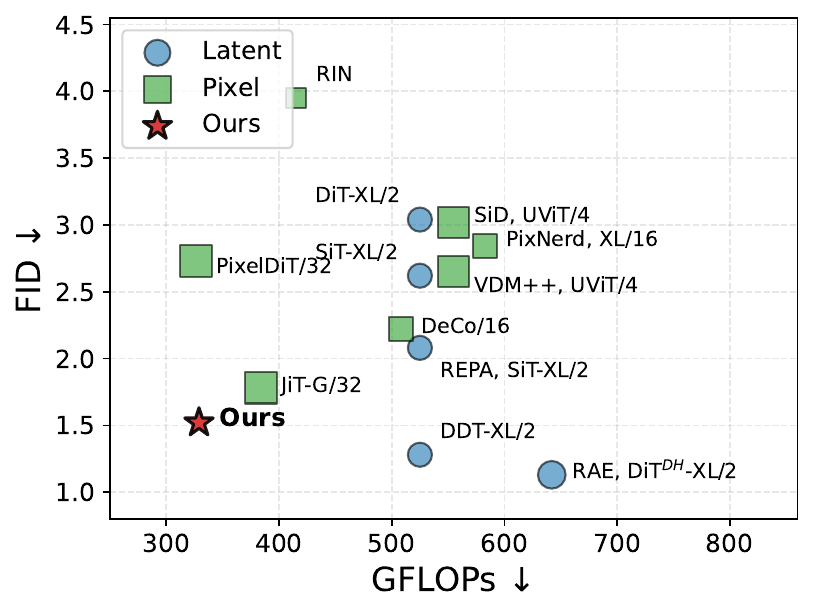}
        \captionof{figure}{Quality--efficiency trade-off on ImageNet $512\times512$. Our method achieves a favorable balance between computational cost and generation quality.}
    \label{fig:tradeoff}
\end{minipage}

\end{figure*}

\begin{figure}[h]
    \centering
    \includegraphics[width=0.95\linewidth]{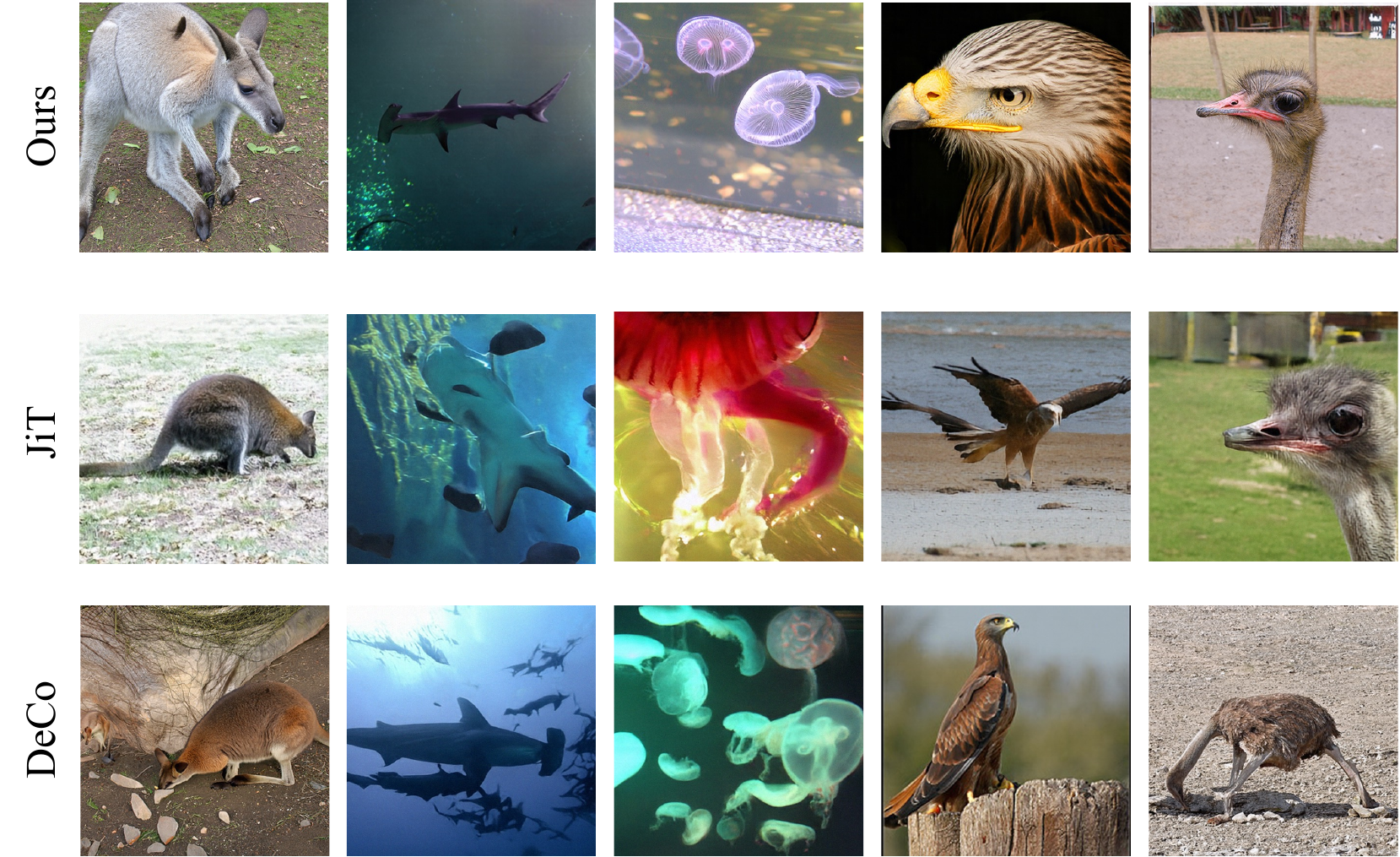}
    \caption{
    \textbf{Qualitative comparison of generated images on ImageNet.} 
    Compared with baseline methods for $512\times512$ resolution, our model produces images with richer local details, sharper structures, and more faithful fine-grained textures. 
    Please zoom in for a better comparison of details.
    }
    \label{fig:image_results}
\end{figure}

\textbf{Implementation details.}
Our training setup closely follows JiT to ensure a fair comparison. Unless otherwise noted, we train our models on ImageNet at $512 \times 512$ resolution and use a patch size of $32$. Following the setting used by JiT, we adopt the $x_0$-prediction parameterization as in JiT, where the model directly predicts the clean image instead of the noise, and set the noise scale to $2.0$. We use Adam~\cite{kingma2014adam} as the optimizer with $\beta_1=0.9$, $\beta_2=0.95$, and a learning rate of $2\times10^{-4}$. Further implementation details are provided in the Appendix.

\paragraph{Better pixel-space generation with lower computation.}
We compare our method with both latent-space and pixel-space diffusion models on ImageNet $512 \times 512$ in Table~\ref{tab:imagenet512_comparison}. Among pixel-space methods, our model achieves the best overall trade-off between computational cost and generation quality. In particular, compared with recent strong pixel-space baselines, our method obtains lower FID while using fewer FLOPs, showing that the proposed dual-branch design improves generation quality without relying on substantially higher computation. This trend is also reflected in Figure~\ref{fig:tradeoff}, where our method achieves a favorable position when jointly considering FID and FLOPs.

Figure~\ref{fig:image_results} provides qualitative comparisons with JiT and other pixel-space diffusion baselines. As shown, our model generates images with richer local details and more faithful fine-grained structures. Baseline models often produce plausible global layouts but tend to smooth out high-frequency textures or miss subtle object details. In contrast, our dual-branch design preserves stronger appearance representations in the pixel branch, enabling better reconstruction of textures, boundaries, and small visual elements. These qualitative results are consistent with the quantitative improvements and support the benefit of separating global structural reasoning from fine-grained appearance modeling.

\begin{table}
\centering
\small
\setlength{\tabcolsep}{6pt}
\begin{threeparttable}
\caption{Comparison with JiT models on ImageNet $512 \times 512$. }
\label{tab:jit_results}
\begin{tabular}{lcccc|lcccc}
\toprule
\textbf{Model} & \textbf{Params} & \textbf{Gflops} & \textbf{FID}$\downarrow$ & \textbf{IS}$\uparrow$ & \textbf{Model} & \textbf{Params} & \textbf{Gflops} & \textbf{FID}$\downarrow$ & \textbf{IS}$\uparrow$\\
\midrule
JiT-B/32~\cite{jit} & 133M & 26  & 4.02 & 271.0  &JiT-H/32~\cite{jit} & 956M & 183 & 1.94 & 309.1 \\
 Ours-B/32 & 132M & 26.8  & 3.75 & 283.3  & Ours-H/32 & 781M & 153 & 1.67 & 311.9\\
JiT-L/32~\cite{jit} & 462M  & 89  & 2.53 & 299.9 &JiT-G/32~\cite{jit} & 2B   & 384 & 1.78 & 306.8\\
 Ours-L/32 & 434M & 86  & 2.22 & 304.7 & Ours-G/32 & 1.69B   & 329 & 1.52 & 317.4\\

\bottomrule
\end{tabular}
\end{threeparttable}
\end{table}

\begin{table}
\centering
\caption{\textbf{Ablation on the size of the extra dimension.} Increasing the dimension improves representation capacity but also increases computational cost. The performance peaks at an intermediate dimension, suggesting that overly large extra dimensions may introduce unnecessary complexity without consistent quality gains.}

\label{tab:ablation_extra_dim}
\small
\setlength{\tabcolsep}{6pt}
\begin{tabular}{c|c|c|c|c}
\toprule
\textbf{Extra Dim} & \textbf{Params (M)} & \textbf{GFLOPs} & \textbf{FID}$\downarrow$ & \textbf{IS}$\uparrow$ \\
\midrule
2048  & 126 & 25.1 & 3.92 & 276.4 \\
4096  & 139 & 28.6 & 3.63 & 289.7 \\
8192  & 165 & 35.8 & \textbf{3.41} & \textbf{304.8} \\
16384 & 218 & 50.3 & 3.57 & 296.2 \\
\midrule
\textbf{Ours (Default, 3072)} & 132 & 26.8 & 3.75 & 283.3 \\
\bottomrule
\end{tabular}
\end{table}

\begin{table}[t]
\centering
\caption{\textbf{Ablation on the information flow in cross-attention. }The default design updates only the subpatch tokens using patch-token context. Updating both branches or introducing subpatch-to-subpatch attention increases computation but leads to worse generation quality.}
\label{tab:ablation_cross_attention_flow}
\small
\setlength{\tabcolsep}{6pt}
\begin{tabular}{c|c|c|c|c}
\toprule
\textbf{Cross-Attention Flow} & \textbf{Params (M)} & \textbf{GFLOPs} & \textbf{FID}$\downarrow$ & \textbf{IS}$\uparrow$ \\
\midrule
Patch $\rightarrow$ Subpatch Only & 132 & 26.8 & \textbf{3.75} & \textbf{283.3} \\
Patch $\leftrightarrow$ Subpatch  & 135 & 28.1 & 4.29 & 276.7 \\
Subpatch $\rightarrow$ Patch Only & 132 & 26.8 & 4.73 & 270.9 \\

Patch $\rightarrow$ Subpatch + Subpatch Self-Attn & 136 & 30.2 & 3.92 & 278.1 \\
\bottomrule
\end{tabular}
\end{table}

\paragraph{Necessity of the additional dimension.}
Table~\ref{tab:jit_results} compares our models with JiT under the same ImageNet $512 \times 512$ setting to examine whether the additional dimension is necessary for pixel-space diffusion transformers. Compared with JiT, our method consistently improves generation quality across all model scales. At the base scale, Ours-B/32 achieves a lower FID of 3.75 compared to 4.02 for JiT-B/32
with almost the same parameter count and GFLOPs. This indicates that the gain does not simply come from increasing model size, but from the proposed additional-dimension design, which provides extra capacity for modeling fine-grained intra-patch appearance.

The advantage becomes more evident at larger scales. Ours-L/32 improves FID from 2.53 to 2.22 while using slightly fewer parameters and FLOPs than JiT-L/32. Similarly, Ours-H/32 and Ours-G/32 outperform their JiT counterparts with substantially lower computational cost. In particular, Ours-G/32 achieves the best FID of 1.52 with 1.69B parameters and 329 GFLOPs, outperforming JiT-G/32, which uses 2B parameters and 384 GFLOPs. These results suggest that explicitly allocating an additional dimension for subpatch-level representation is an effective and efficient way to overcome the representation bottleneck of standard pixel transformers. Rather than relying solely on scaling the backbone, our design improves the quality-efficiency trade-off by separating global patch-level reasoning from fine-grained local appearance modeling.

\paragraph{Effect of the extra dimension size.}
Table~\ref{tab:ablation_extra_dim} studies the influence of the extra dimension size in our dual-branch pixel modeling framework. Starting from a smaller extra dimension, increasing the dimension consistently improves generation quality, reducing FID from 3.92 at 2048 dimensions to 3.41 at 8192 dimensions, while improving IS from 276.4 to 304.8. This indicates that a larger extra branch provides stronger capacity to preserve fine-grained intra-patch appearance, thereby alleviating the representation bottleneck caused by compressing each raw patch into a compact token. However, further increasing the dimension to 16384 yields no additional gains and slightly degrades both FID and IS, despite substantially increasing the parameter count and computational cost. This suggests that overly large extra dimensions may introduce optimization difficulty. We therefore use 3072 as the default setting, which offers a favorable balance between efficiency and generation quality.

\paragraph{Ablation on cross-attention flow.}
Table~\ref{tab:ablation_cross_attention_flow} evaluates different information-flow designs in the cross-attention module. Our default design only updates the subpatch tokens by attending to the globally contextualized patch tokens. This asymmetric design achieves the best FID and IS, indicating that the compact patch branch is most effective when used as a stable structural bottleneck, while the high-dimensional subpatch branch absorbs global context for appearance refinement. In contrast, bidirectional interaction degrades performance despite additional computation, suggesting that feeding local subpatch information back into the patch branch may disturb its global representation. We also find that introducing subpatch-to-subpatch attention does not improve performance and instead slightly worsens generation quality. This implies that explicit dense interaction among subpatch tokens is unnecessary in the cross-attention module, as local appearance modeling is better handled through structured subpatch representations conditioned on patch-level context. These results validate the asymmetric patch-to-subpatch cross-attention used in our default architecture.

\paragraph{Ablation on projection and modulation design.}
Table~\ref{tab:ablation_modulation} studies different projection and modulation mechanisms for integrating the extra dimensions. Compared with using a single shared small MLP for all subpatch tokens, our default per-token projection achieves substantially better generation quality.
This suggests that different subpatch tokens encode distinct local appearance structures and should not be forced to share the same projection function. We also compare against a large MLP that first concatenates all subpatch tokens and then projects them jointly. Although this design introduces more parameters and computation, it performs worse than the lightweight per-token design, indicating that simply increasing projection capacity does not necessarily lead to better subpatch modeling and may make optimization more difficult. We further evaluate the modulation strategy used for different subpatch tokens. Applying the same modulation to all subpatch tokens leads to slightly worse performance than using different modulation parameters, suggesting that token-specific modulation provides a more flexible way to adapt local subpatch representations.

\begin{table}
\centering
\caption{
Ablation on projection and modulation designs for integrating the extra dimensions. 
The default setting uses a small per-token MLP together with different modulation parameters.
}
\small
\setlength{\tabcolsep}{6pt}
\begin{tabular}{l|c|c|c|c}
\toprule
\textbf{Variant} & \textbf{Params (M)} & \textbf{GFLOPs} & \textbf{FID}$\downarrow$ & \textbf{IS}$\uparrow$ \\
\midrule
\multicolumn{5}{l}{\textit{Projection design}} \\
Small MLP (single)        & 128 & 25.7 & 4.18 & 267.9 \\
Large MLP                 & 156 & 31.4 & 3.91 & 276.8 \\
\textbf{Small MLP (per token)} & \textbf{132} & \textbf{26.8} & \textbf{3.75} & \textbf{283.3} \\
\midrule
\multicolumn{5}{l}{\textit{Modulation design}} \\
Same Modulation           & 131 & 26.5 & 3.96 & 278.6 \\
\textbf{Different Modulation}  & \textbf{132} & \textbf{26.8} & \textbf{3.75} & \textbf{283.3} \\
\bottomrule
\end{tabular}

\label{tab:ablation_modulation}
\end{table}

\section{Discussion, Limitations, and Future Work}

Our results show that explicitly separating global structural reasoning from fine-grained appearance modeling is beneficial for pixel-space diffusion transformers. Rather than uniformly increasing the entire model, our design allocates additional capacity to the pixel branch, where high-frequency details are most needed, while keeping the global reasoning branch relatively compact.

The main limitation of our approach is the additional training cost. The high-dimensional pixel branch increases memory usage, particularly for storing intermediate activations, and makes the backward pass more expensive due to the extra projection, modulation, and cross-branch interaction modules. As a result, training larger variants requires more GPU memory and longer training time than a standard single-branch transformer. Future work includes improving the efficiency and scalability of this architecture. Possible directions include memory-efficient training strategies, sparse or local cross-branch interactions, parameter sharing across subpatch groups, and adaptive allocation of the extra dimensions. Another promising direction is to extend the proposed dual-branch design to video generation, where global spatio-temporal structure and local appearance details are both critical for high-quality synthesis.
\newpage
\bibliography{ref}
\bibliographystyle{plain}

\end{document}